\documentclass[11pt]{article}

\usepackage[a4paper,left=15mm,top=20mm,right=15mm,bottom=25mm]{geometry}

\usepackage{graphicx}

\usepackage{amssymb}
\usepackage{amsthm, amsmath, amsfonts}
\usepackage{mathtools}
\usepackage{mathptmx}
\usepackage[mathscr]{euscript}

\usepackage{xcolor}
\usepackage[colorlinks]{hyperref}
\hypersetup{
	citecolor=blue,
	linkcolor=blue,
	urlcolor=blue
}

\usepackage[font=footnotesize]{caption}
\usepackage{subcaption}
\usepackage{authblk}

\makeatletter
\def\moverlay{\mathpalette\mov@rlay}
\def\mov@rlay#1#2{\leavevmode\vtop{%
		\baselineskip\z@skip \lineskiplimit-\maxdimen
		\ialign{\hfil$\m@th#1##$\hfil\cr#2\crcr}}}
\newcommand{\charfusion}[3][\mathord]{
	#1{\ifx#1\mathop\vphantom{#2}\fi
		\mathpalette\mov@rlay{#2\cr#3}
	}
	\ifx#1\mathop\expandafter\displaylimits\fi}
\makeatother

\providecommand{\keywords}[1]{\textbf{\textit{Keywords: }} #1}

\title{Automatic Bounding Box Annotation with Small Training Data Sets for Industrial Manufacturing}

\author[1]{Manuela Gei{\ss}}
\author[1]{Raphael Wagner}
\author[2]{Martin Baresch}
\author[2]{Josef Steiner}
\author[1]{Michael Zwick}

\affil[1]{Software Competence Center Hagenberg GmbH, Softwarepark 32a, 4232 Hagenberg, Austria}
\affil[2]{KEBA Group AG, Reindlstra{\ss}e 51, 4040 Linz, Austria}

\date{\ }

\begin{document}
\sloppy

\maketitle

\abstract{
In the past few years, object detection has attracted a lot of attention in the context of human-robot collaboration and Industry 5.0 due to enormous quality improvements in deep learning technologies. In many applications, object detection models have to be able to quickly adapt to a changing environment, i.e., to learn new objects. A crucial but challenging prerequisite for this is the automatic generation of new training data which currently still limits the broad application of object detection methods in industrial manufacturing.
In this work, we discuss how to adapt state-of-the-art object detection methods for the task of automatic bounding box annotation for the use case where
the background is homogeneous and the object’s label is provided by a human. We compare an adapted version of Faster R-CNN and the Scaled Yolov4-p5 architecture and show that both can be trained to distinguish
unknown objects from a complex but homogeneous background using only a
small amount of training data.
}

\bigskip
\noindent


\section{Introduction}

For small-lot industrial manufacturing, reconfigurability and adaptability  becomes a major factory in the era of Smart Factories and Industry 5.0. \cite{Mehrabi2000ReconfManufacturing}. Especially in the context of human-robot collaboration (Cobots, \cite{matheson2019Cobot}; Industry 5.0), a robot needs to be able to quickly adapt to changing demands of a human operator. In such a setting, the ability to detect objects in the working area is crucial for a robot to react to an operator's commands. State-of-the-art (deep learning based) object detection models usually operate under a closed-world assumption, i.e., all objects required for detection are known beforehand and included in the training data set. However, a robot that has to react to a changing working environment, is often confronted with an open-world setting (open-set/world recognition, \cite{scheirer2012toward,bendale2016towards,bendale2015towards}), where new objects not present in the training data set during model training need to be recognized. Besides the ability for fast retraining of the object detection model \cite{geissdexa2022}, the question of how much data is needed for training is of great importance in a challenging environment such as industrial manufacturing \cite{Lee2020PitfallsAP}. Notably automatic labeling of new training data is essential for a robot to adapt to changing environments, in order to avoid the usually time-consuming and expensive manual labeling process.

To this end, we propose a deep learning based automatic labeling approach for fast retraining of object detection models in human-robot collaboration settings for smart factories. In recent years, significant efforts have been made to automate all aspects of the traditional machine learning pipeline (AutoML, \cite{HE2021106622}). In object detection, where manually labeling bounding boxes is especially time consuming, the potential benefits of an AutoML-based approach are particularly high and a crucial factor in enabling human-robot collaboration. In our use case we assume a homogeneous background (but not necessarily simple, see Fig.~\ref{fig:iCub_anno}), where new objects are incrementally learned with a human operator simply initiating the retraining while only providing the new class label. We show that classical state-of-the-art object detection models, such as Faster R-CNN and Yolo, can be used to distinguish new objects from such a homogeneous background and that a small data set is sufficient for training the model for this task. Using object detection approaches in our setting has the added advantage that a single model architecture can be used for both tasks, bounding box annotation and object detection.

Our work is structured as follows: Sec.~\ref{sec:related} gives a short overview of deep learning based object detection methods relevant to our approach and the current state-of-the-art in automated data annotation. Sec.~\ref{sec:appr} describes the use case and test setting that our approach was deployed on, the overall workflow, as well as the object detection architectures used during analysis. Sec.~\ref{sec:experiments} gives an overview of the data sets used (one specifically generated for this work, one publicly available) and details the results of our experiments. Finally, Sec.~\ref{sec:summary} highlights our key findings and discusses potential further improvements to our work.

\section{Related Work}
\label{sec:related}

\subsection{Object Detection}
Deep learning has been a driving force in the field of machine learning that revolutionized many tasks  such as object detection \cite{lecun2015deep}, i.e. the task of classifying and localizing objects in images. Object detection methods are utilized in various applications such as face recognition \cite{jiang2017face}, self-driving cars \cite{Simon_2019_CVPR_Workshops}, and fruit recognition in the context of robotic harvesting \cite{bargoti2017deep}. 
Although the history of multilayer networks dates back to the middle of the 20th century, deep learning has only recently become popular with the development of  high performance parallel computing (e.g. GPU clusters) and the availability of  large annotated data sets such as ImageNet \cite{russakovsky2015imagenet} for training large network structures \cite{Zhao2019}. One important breakthrough  that marked a milestone for the wide application of deep learning methods and laid the ground for modern object detection methods,  was the development of  AlexNet \cite{krizhevsky2012imagenet}, a deep Convolutional Neural Network (CNN) that achieved outstanding results in the popular ImageNet Challenge (ILSVRC) \cite{russakovsky2015imagenet} in 2012. During the last decade, many CNN-based methods for object detection have been developed, which can be mainly categorized into two classes: the two-step region proposal based methods such as Faster R-CNN \cite{ren2015faster}, FPN \cite{Lin_2017_CVPR}, and Mask R-CNN \cite{He_2017_ICCV}, as well as the one-step anchor box based approaches such as the Yolo-family \cite{Redmon_2016_CVPR,bochkovskiy2020yolov4,wang2021scaled} and SSD \cite{Wei2016} (see e.g. \cite{Zhao2019} for a detailed review). These two classes differ in their accuracy-speed trade-off with anchor box based methods having much smaller inference time, while region proposal based achieving higher accuracy \cite{huang2017speed,lin2017focal}. However, recent results suggest that newer versions of Yolo can attain accuracy levels comparable to region proposal based networks \cite{kim2020comparison,yolox2021}.

\subsection{Annotation}
Current object detection methods fall within  the category of supervised learning algorithms with complex network architectures that typically include millions of learnable parameters and therefore require large labeled data sets for training \cite{lecun2015deep,Zhao2019}. The most popular open, large-scale data sets for object detection are the ImageNet \cite{russakovsky2015imagenet}, Pascal VOC \cite{everingham2010pascal}, and MS COCO \cite{lin2014microsoft} data sets, all containing thousands of annotated images per class. However, the annotation of images is a time-consuming and therefore costly task. 
This raises the need for solutions for automated annotation of images. While the problem of automatic image annotation for classification tasks has been treated for more than two decades now (see e.g.\ \cite{cheng2018survey} for a detailed review), the more challenging task of bounding box annotation has only come into focus in the last few years. 
The available methods for the latter task range from inferring the location of object proposals from edges \cite{papadopoulos2016we}, using predictions from a U-Net neural network as a basis \cite{wu2020automatic}, and training the detector model on a subset of manually labeled images \cite{adhikari2018faster}. Such methods have been used in various application such as industrial visual inspection \cite{ge2020towards}, radiology \cite{wu2020automatic}, 3D images \cite{kiyokawa2019fully,apud2021automated}, and object tracking in videos \cite{le2020toward}.
However, the existing approaches typically rely on some human intervention to tackle the trade-off between accurate, manually labeled training data and prediction accuracy of the trained models. In most cases, this intervention is some sort of verification \cite{papadopoulos2016we,konyushkova2018learning} or manual correction \cite{adhikari2021iterative} of the inferred bounding boxes. 
Another strategy is active learning \cite{10.1145/3472291} which reduces the annotation costs by sampling the most informative unlabeled images that are then labeled by a human operator.  This decreases the amount of manual labeling but is usually computationally expensive due to the sampling step. The related assistive learning workflow of \cite{wong2019assistive} reduces the computational costs by including some contextual information in the sampling method, however, it still requires a human-in-the-loop.
Furthermore, there exist labeling approaches using attention maps to extract bounding boxes, however the accuracy is limited and a manual control step required \cite{wu2020automatic}.

For our use case of continuously learning new objects in a short time, manual correction steps are not possible due to time restrictions. On the other hand, we have seen in previous experiments during our project that, for good training results, the bounding boxes do not have to be perfectly accurate in the sense that some pixels can be missed in some cases. Our approach of training object detector models to distinguish between objects and background does meet these requirements.

\section{Approach}
\label{sec:appr}

One of the main problems of annotating images for object detection tasks is the correct identification of the bounding box. In this work, we consider the use case of a \emph{homogeneous} background, which allows us to use state-of-the art object detection methods to distinguish the object from the background.

	\subsection{Use Case}
\begin{figure}[htb]
	\begin{tabular}{lcr}
		\begin{minipage}{0.55\textwidth}
			\centering
			\includegraphics[width=0.8\linewidth]{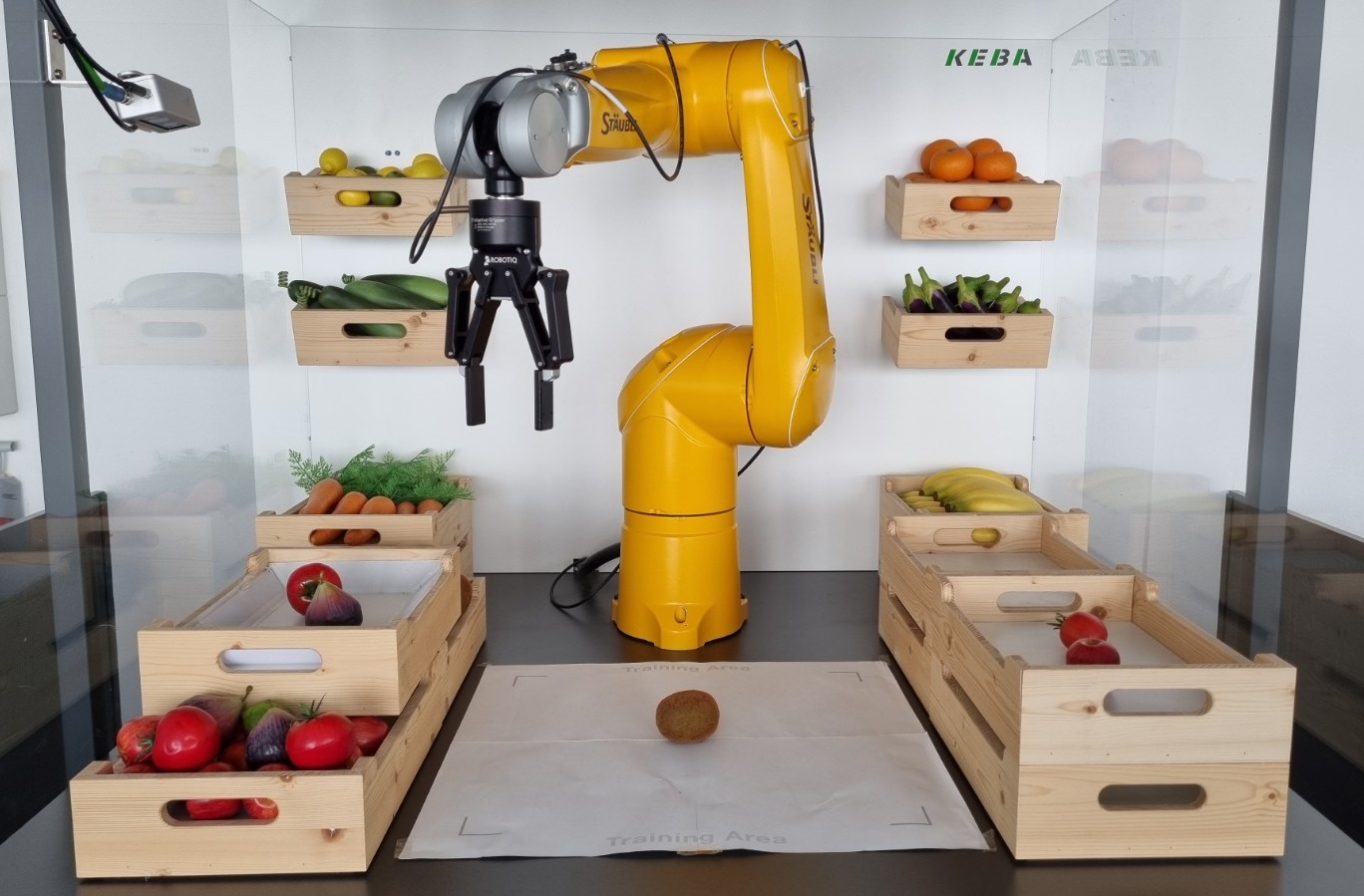}
		\end{minipage}
		& 
		\begin{minipage}{0.35\textwidth}
			\caption{Showcase to demonstrate fast retraining of new object classes: A robot sorting different types of fruits. The left box holds a mixture of different fruits, from which the robot picks up specific ones and transfers them to the correct basket on the right. The new object needs to be placed in the center of the white area by a human operator for the generation of training images. These images are taken by a camera at the robot's gripper while the robot is moving around the object.}
			\label{fig:usecase}
		\end{minipage}
	\end{tabular}
\end{figure}

As a use case we consider a robot that has to recognize objects and sort them into predefined boxes (see setup in Fig.~\ref{fig:usecase}). In addition, our setup is that of an open world, that is, the robot has to learn new objects at later time points without forgetting the old ones. For this object detection task, we assume that a human operator initiates the training of the new object by presenting the object to a camera and providing the name of the object. The rest of the pipeline, that is the generation of training images, labeling of the images and training of the model, is then performed automatically. In this work, we focus on  the automatic annotation of the training and validation images. For a description of the whole pipeline including automatic image generation and the training process, we refer to our recent contribution \cite{geissdexa2022}.


The automatic annotation of images for object detection is in general not an easy task. First, it is often not known at which time point a new object class has to be learned and what the label of the new object class is. The identification of an object to be unknown is a non-trivial task, which is not inherently captured by the design of current deep learning models \cite{bendale2016towards}. 
Second, for retrieving the corresponding bounding box, the model has to distinguish a new object from some - often complex and previously unknown - background. 
Our use case differs from general object detection tasks in the following main points: (i) there is a human signal whenever a new object has to be learned, (ii) the name of the new object class is given to the AI model, (iii) the new class is trained on images containing only a single object, and (iv) the background remains approximately the same, i.e., we can speak of a \emph{homogeneous} background. By \emph{homogeneous} we not necessarily mean that the background must be monochromatic or exactly the same in each image (e.g.\ a white background in all images) but we also speak of an homogeneous background if, for instance, each image contains different parts of a possibly complex environment (see, e.g.,\ the iCubWorld data set in Sec.~\ref{sec:data}).

\subsection{Our Approach}
\label{sec:appproach}
The task of annotating an image for object detection consists of two separate subtasks:  (1) determining the class label and (2) finding the bounding box. In our use case, the image that needs to be annotated contains exactly one object and the label of its class is provided by the human operator. Thus, the class label  is already given and it remains to find the corresponding bounding box coordinates.
For this task, our approach is to train some state-of-the-art object detection method for background separation and distinguish the new object from the background. As training data, we provide the network with annotated images of different objects in the same homogeneous background setup. In contrast to conventional training of object detection models, in our approach the same label ``object'' is given to each object in the training and validation data. After successful training, the bounding box coordinates of the object in the test image can now be extracted by applying the trained object detection model on the image and retrieving the bounding box of the resulting prediction. For some models, we found it to be beneficial to include two additional post-processing steps to further refine the bounding box (for examples also see Fig.~\ref{fig:frcnn_inference}):
\begin{itemize}
	\item[(P1)] if the model erroneously predicted more than one bounding box per image, merge all bounding boxes into one, which is the smallest bounding box containing all others,  
	\item[(P2)] add some additional slack, that is, increase the bounding box by a few pixels on each side.
\end{itemize}

This workflow will be applied to two different state-of-the-art object detection networks, the two-step Faster R-CNN model and one representative from the one-step Yolo-family.

\subsubsection{Faster R-CNN}
\begin{figure*}[htb]
	\linespread{0.7}
	\includegraphics[trim=0cm 5cm 0cm 6cm, clip, width=\textwidth]{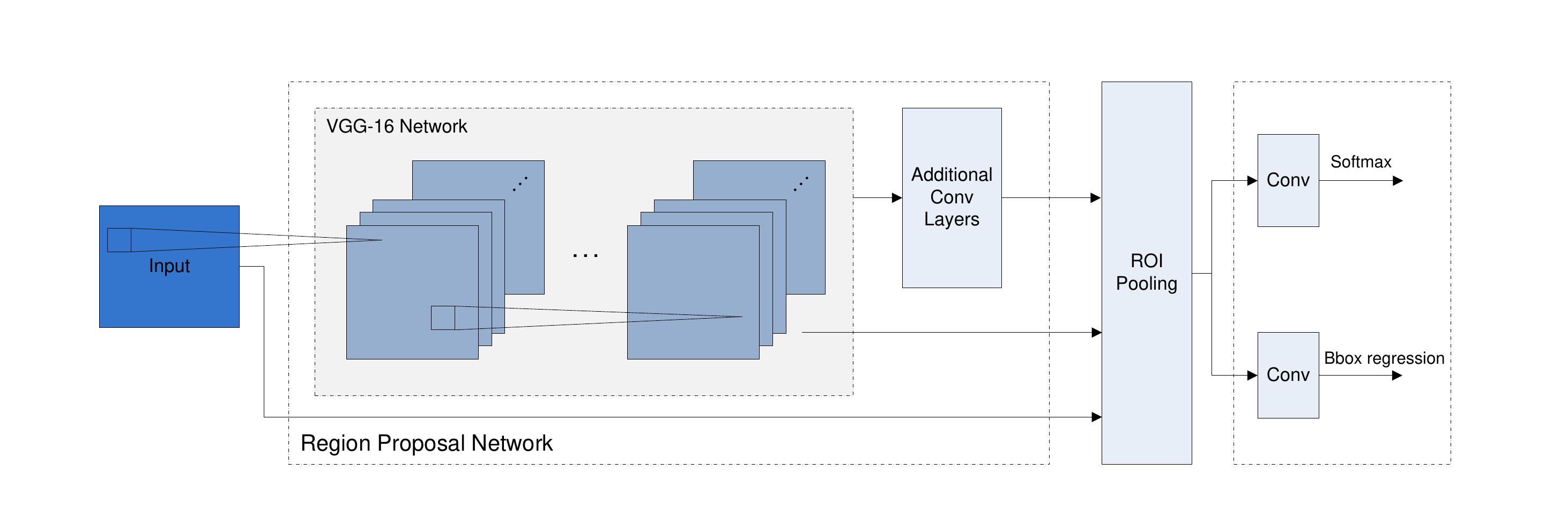}
	\captionsetup{singlelinecheck=false, width=0.9\linewidth}
	\caption{Architecture of the used Faster R-CNN model with a VGG-16 backbone \cite{geissdexa2022}.}
	\label{fig:frcnn}
\end{figure*}

The architecture of the Faster R-CNN network consists of two stages:  (i) a region-proposal network with a feature extractor network (in our case this is the VGG-16 backbone) followed by additional convolutional layers, and (ii) the network's head consisting of two outputs, one for object classification and the other for bounding box regression. These two parts of the network are connected by an ROI Pooling Layer. The second part of the network typically consists of fully connected layers. However, due to some technical limitations in the overall project of our use case (see below), we replaced these fully connected layers by convolutional layers, more precisely, our network's head consists of one single convolutional layer with Softmax activation for  the classification and one convolutional layer with linear activation for bounding box regression. These technical limitations result from running the inference of our trained Faster R-CNN network on an FPGA using the Deep Neural Network Development Kit (DNNDK)\footnote{\url{https://www.xilinx.com/support/documentation/user_guides/ug1327-dnndk-user-guide.pdf}} from Xilinx (see also \cite{geissdexa2022} for a description of our full pipeline). The architecture is shown in Fig.~\ref{fig:frcnn}.

\subsubsection{Yolov4-p5}
\begin{figure*}[htb]
	\linespread{0.7}
	\centering
	\includegraphics[width=0.9\textwidth]{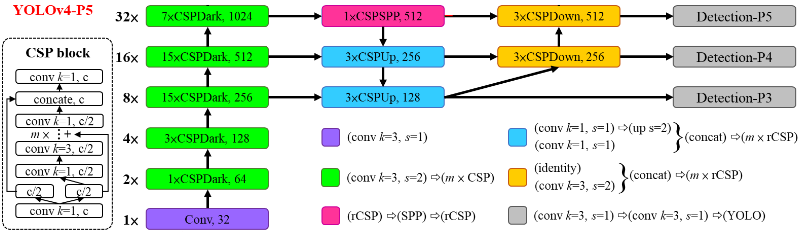}
	\captionsetup{singlelinecheck=false, width=0.9\linewidth}
	\caption{The architecture of Yolov4-p5. Figure adapted from \cite{wang2021scaled}.}
	\label{fig:yolo}
\end{figure*}

From the YOLO-family of model architectures, we chose the recently developed Scaled-Yolov4-p5 \cite{wang2021scaled}. The framework is based on the large branch of the official pytorch implementation\footnote{\url{https://github.com/WongKinYiu/ScaledYOLOv4}}. The architecture of the Scaled-Yolov4-p5 model is summarized in Fig.~\ref{fig:yolo}. It consists of 32 modules in total with 476 layers containing about 70 million parameters. 

\section{Experiments}
\label{sec:experiments}

\subsection{Data Sets}
\label{sec:data}
The workflow presented in the previous section has been applied to two different data sets with a homogeneous background (summarized in Tab.\ \ref{tab:models}). 

\subsubsection{Fruits}	
\begin{figure}[htb]
	\begin{tabular}{cc}
		\begin{minipage}{0.5\textwidth}
			\centering
			\includegraphics[width=0.7\linewidth]{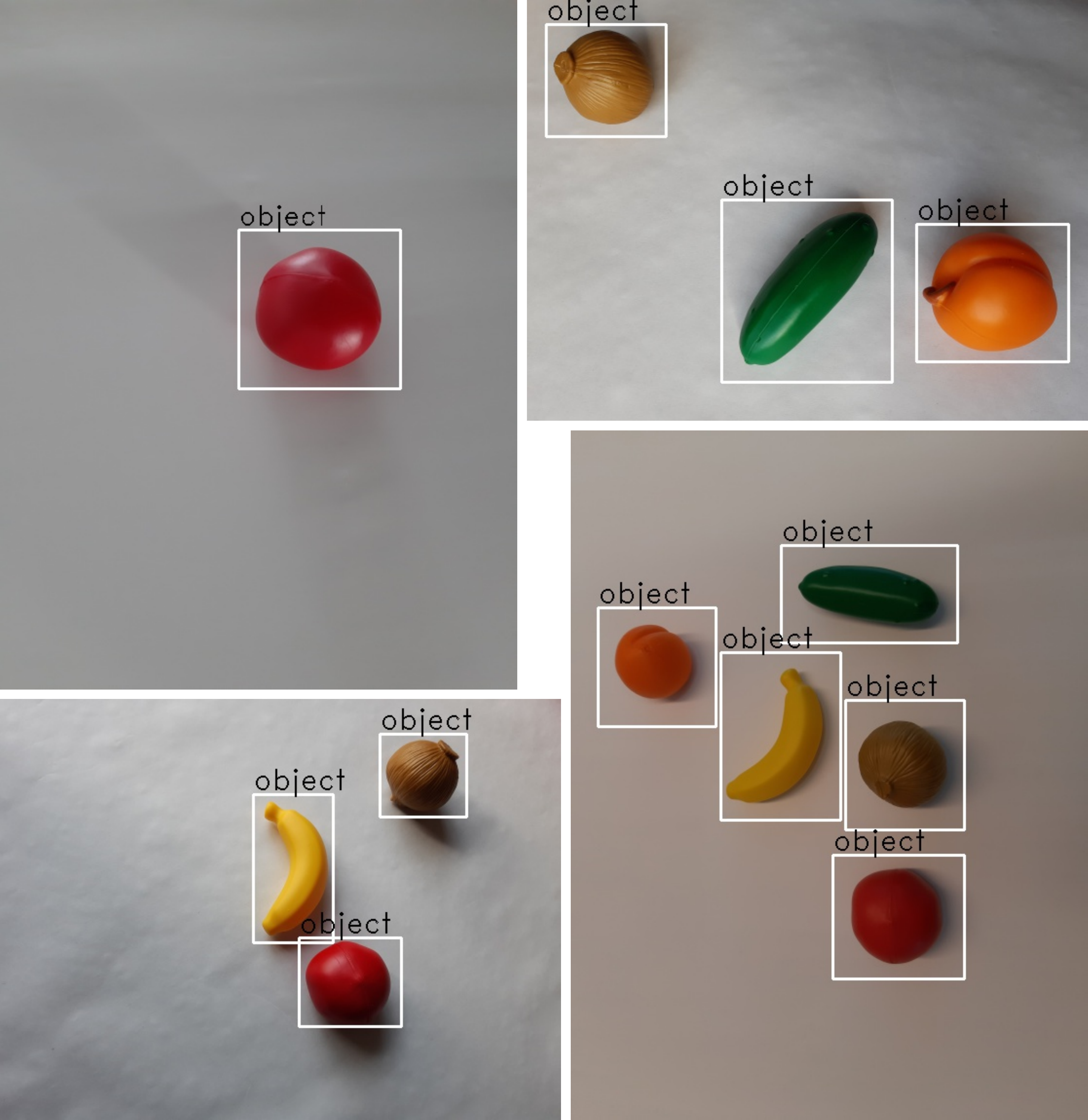}
		\end{minipage}
	 &
		\begin{minipage}{0.35\textwidth}
			\caption{Annotation examples from our fruits data set. Annotations have been generated manually using LabelImg. A piece of paper serves as background, however the lightning differs and often induces shadows.}
			\label{fig:fruits_data}
		\end{minipage}
	\end{tabular}
\end{figure}

This data set has been created during the course of one of our industrial projects. It consists of 330 images of five different imitation fruits and vegetables made of plastic (apricot, banana, cucumber, onion, tomato) on a white background. Approximately half of these images contain only one object, the rest contains a mix of different object classes with up to four objects. The pictures are of size 708x531 and 576x432 resp.\ 531x708 and 432x576, and have been manually annotated using the labeling tool LabelImg\footnote{\url{https://github.com/tzutalin/labelImg}} (see Fig.~\ref{fig:fruits_data} for some examples). This data is split into training and validation data sets (approximately 90\% resp.\ 10\% of the images). As a test data set we use 100 images of size 800x600, each containing exactly one object on a white background. This data set contains 20 different classes (5 images each) including the training objects and new objects such as other fruits and vegetables, pliers, measuring tape etc. (see Fig.~\ref{fig:frcnn_inference}).

\subsubsection{iCubWorld}	
\begin{figure*}[htb]
	\centering
	\includegraphics[width=0.9\textwidth]{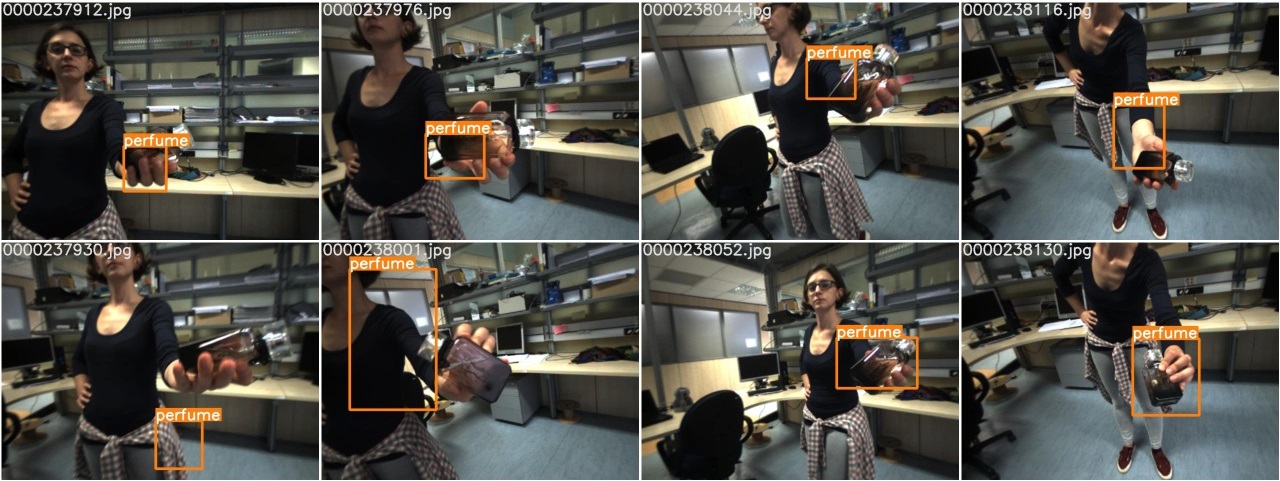}
	\captionsetup{singlelinecheck=false, width=0.9\linewidth}
	\caption{Example images from the iCubWorld data set including the ground truth annotations that have been automatically generated from a human-robot-interaction. In many cases the object has not or only inaccurately been localized.}
	\label{fig:iCub_anno}
\end{figure*}

As a second data set we chose the iCubWorld\footnote{\url{https://robotology.github.io/iCubWorld/}} data set \cite{fanello2013icub}. It contains more than 400k images of 20 different classes. 
In contrast to the fruit data set, the background in the iCubWorld data set is not exactly the same in all images, but homogeneous in the sense that all pictures have been taken in the same environment, that is at different positions in one university lab and with the object held by the same person with changing clothing. The data contains annotation information for object detection, where the annotations also contain information about different poses such as ``Mix'', ``2D rotation'' etc., describing how the images change from one frame to the next. In the original data set, these annotations have been automatically generated by a robot during a human-robot-interaction, where the human provides the label verbally and shows the object in their hand. The robot then localizes the object by tracking either motion or depth information. Example images including the ground truth annotations are shown in Fig.~\ref{fig:iCub_anno}. It can be seen that the annotations are not optimal in many cases. For this reason, we again used LabelImg for manually creating our own ground truth annotations. We randomly chose 659 images of seven object classes (mug, pencil case, ring binder, soap dispenser, soda bottle, squeezer, sunglasses) with 83-105 images per class and split this into training and validation data set (90\% resp.\ 10\% of the images). As a test data set we use a different subset of 82 images that contains all 20 object classes. 

\subsection{Experiments and Results}
 
\subsubsection{Faster R-CNN}
\label{sec:results_frcnn}
For the experiments with our adapted Faster R-CNN architecture (see Fig.~\ref{fig:frcnn}), we used the pretrained weights from the Keras Applications Module\footnote{\url{https://github.com/fchollet/deep learning-models/releases}} for the VGG-16 backbone. We then trained two models: the model $F$ on the fruits data set and the model $I$ on the iCubWorld data set (see Tab.\ \ref{tab:models}). Both models have been trained for 200 epochs without freezing layers. We also tried freezing the VGG-16 backbone. However, the results were not satisfying, therefore we do not further discuss this here.

\begin{table*} [htb] 
	\centering   
    \begin{subtable}[c]{0.9\linewidth}
    \begin{tabular}{c|cc|cc|c} 
      \textbf{Data Set} & \multicolumn{2}{|c|}{\textbf{No.\ images}} & \multicolumn{2}{|c|}{\textbf{No.\ classes}} & \textbf{Training classes contained}\\
      & train \& val & test & train \& val & test & \textbf{in test data}\\
      \hline
      Fruits & 330 & 100 & 5 & 20 & \emph{Some} contained\\
      iCubWorld & 659 & 82 & 7 & 20 & \emph{All} contained\\
    \end{tabular}
	\captionsetup{singlelinecheck=false}
    \caption{Data sets used in our study. }
    \end{subtable}
    
    \vspace{0.4cm}
    
    \begin{subtable}[c]{0.9\linewidth}
    \begin{tabular}{c|c|c} 
      \textbf{Model} & \textbf{Training data set} & \textbf{Post-processing (P1) \& (P2)}\\
      \hline
      $F$ & Fruits & no\\
      $F^+$ & Fruits & yes\\
      $I$ & iCubWorld & no\\
      $I^+$ & iCubWorld & yes\\
      \end{tabular}
  	\captionsetup{singlelinecheck=false}
     \caption{Models used in our study.}
    \end{subtable}
	\captionsetup{singlelinecheck=false, width=0.9\linewidth}
    \caption{Summary of (a) data sets and (b) models used in our study. Training and validation images are splitted into 90\% training and 10\% validation data.}
    \label{tab:models}
\end{table*}

\begin{figure*}[h!]
	\begin{minipage}[c]{1.0\linewidth}
		\subcaption[l]{\textbf{Metrics}}
		\centering
		\includegraphics[width=0.9\textwidth]{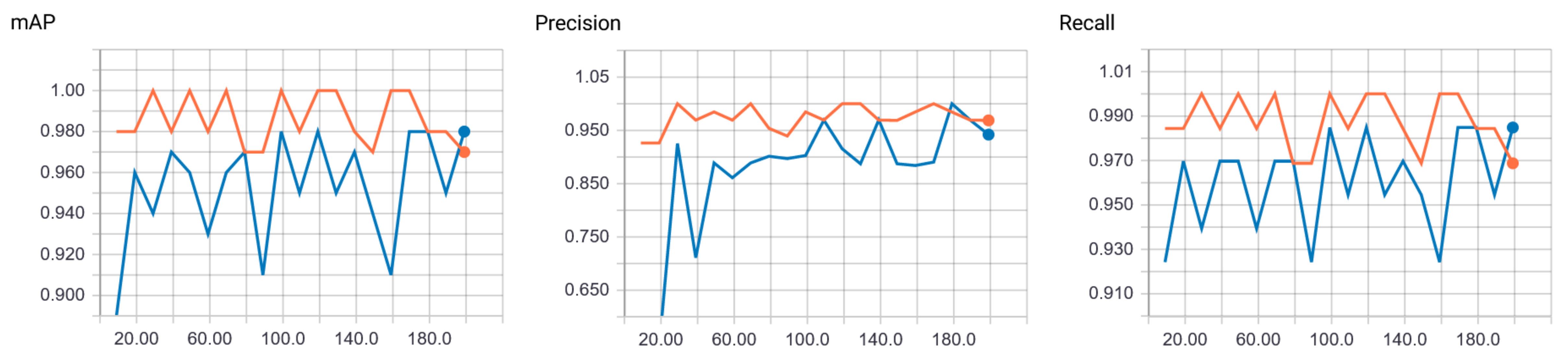}
		\label{fig:frcnn_metrics}
	\end{minipage} 
	
	\begin{minipage}[c]{1.0\linewidth}
		\vspace{0.2cm}
		\subcaption[l]{\textbf{Inference examples: Fruits model $F$ and $F^+$}}
		\centering
		\includegraphics[width=0.9\textwidth]{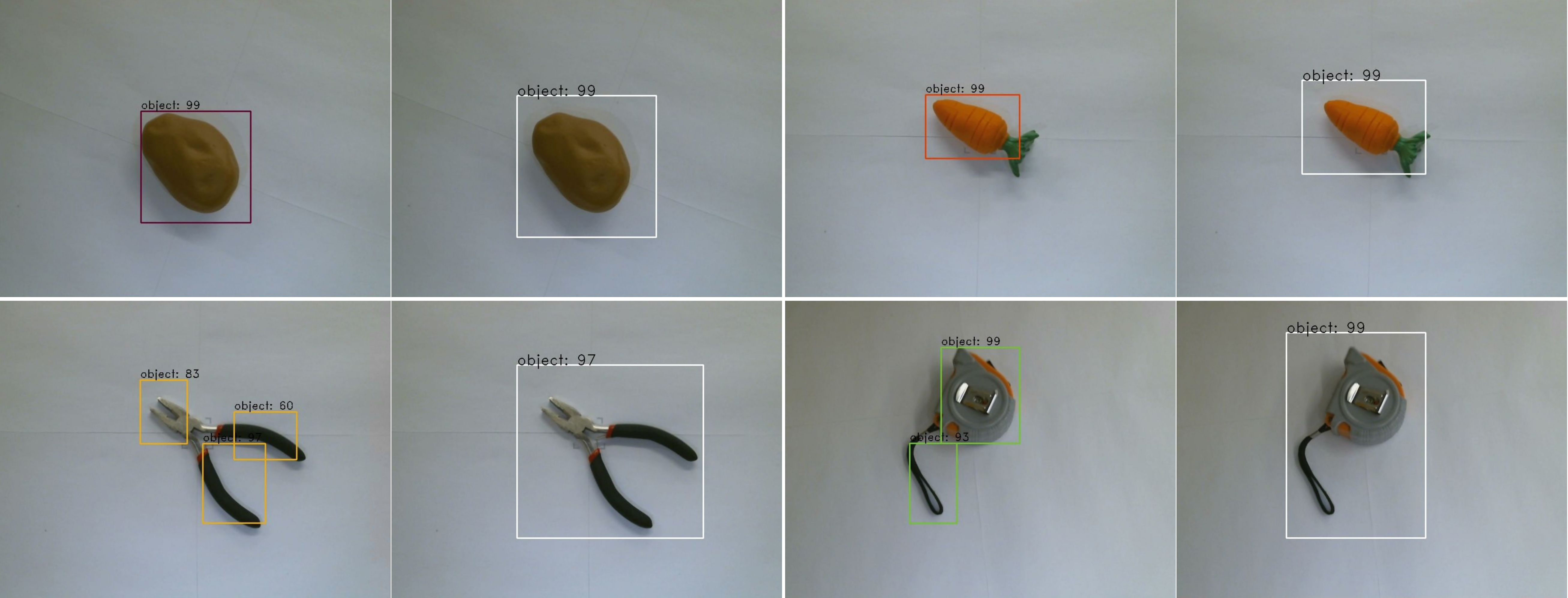}
		\label{fig:frcnn_inference}
	\end{minipage} 
	
	\begin{minipage}[c]{1.0\linewidth}
		\vspace{0.2cm}
		\subcaption[l]{\textbf{Inference examples: iCubWorld model $I$}}
		\centering
		\includegraphics[width=0.9\textwidth]{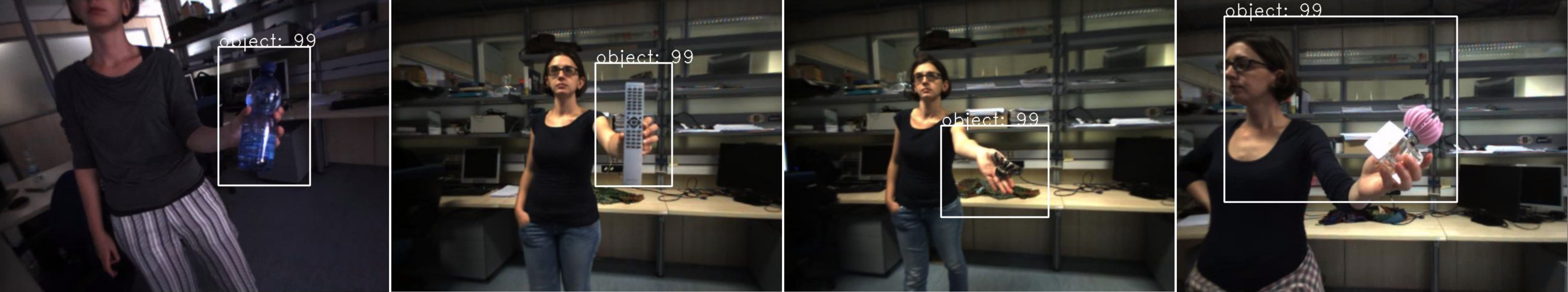}
		\label{fig:icub_inference}
	\end{minipage}
	\captionsetup{singlelinecheck=false, width=0.9\linewidth}
	\caption{Results on the fruits data set and the iCubWorld data set using Faster R-CNN. For both, the corresponding model $F$ resp.\ $I$ has been trained for 200 epochs.
			\textit{Panel (a):} The metrics (mAP@0.5, recall and precision) on the validation data sets are calculated after each $10^{th}$ epoch (orange: fruits, blue: iCubWorld). For both models $F$ and $I$, the metrics reach high values after only a few epochs.
			\textit{Panel (b):} Inference examples of $F$ and $F^+$ on the fruits test data set. For each object, the result of $F$ resp.\ $F^+$ is shown on the left resp.\ right. The potato is well annotated by both models, while the carrot benefits from the additional slack in $F^+$. For the pliers and the tape, $F$ detects multiple bounding boxes resulting in a very good annotation in $F^+$. 
			\textit{Panel (c):} Inference examples of $I^+$ on the iCubWorld test data set. The first two examples are well annotated, while the inferred bounding boxes in the remaining examples are too large due to inclusion of the human hand (hair clip) and incorrectly detected boxes merged with correct boxes (perfume).\\
			More detailed results on the absolute numbers of correct and incorrect annotations can be found in Fig.~\ref{fig:results_frcnn_more}.
			}
	\label{fig:results_frcnn}
\end{figure*}

The metrics (mAP@0.5, recall, precision) have been calculated after each $10^{th}$ epoch and reach very high values after only a few epochs (see Fig.~\ref{fig:frcnn_metrics}). Model $F$ performs slightly better than model $I$, where $F$ reaches mAP@0.5 values between 0.97 and 1.0 during the entire training period, starting with 0.98 at epoch 10. The lower performance of model $I$ is not surprising due to the more complex background in  the images of the iCubWorld data set. However, $I$ also shows  good performance with a mAP@0.5 constantly higher than 0.91 after only 20 epochs. 
Inference examples of $F$ on test data (see Fig.~\ref{fig:frcnn_inference}) show that (i) the predicted bounding box is often tight around the object with sometimes missing parts of the object (see e.g.\ carrot), and (ii) in some cases the object is not detected by one single bounding box but several  boxes cover different parts of the object. These multiple bounding boxes particularly occur for objects whose form is very dissimilar to the forms of the objects in the training data set (e.g.\ pliers and tape with long, almost separated parts). In order to reduce such errors, we applied the two post-processing steps (P1) and (P2) introduced in Sec.~\ref{sec:appproach} to merge multiple bounding boxes and enlarge the boxes by some additional slack (15 resp.\ 10 pixels for the fruits  resp.\ iCubWorld data set). The resulting models are denoted by $F^+$ resp.\ $I^+$ for the fruits resp.\ the iCubWorld data set. Although the performance of model $F$ is high, even after only a short amount of training time, these post-processing steps further increase the number of good annotations from 32 to 84 out of 100 test images at epoch 10 and from 51 to 87 at epoch 200 (see Fig.~\ref{fig:results_frcnn_more} in the appendix for a detailed list of the absolute numbers). The remaining 12 cases in epoch 200 of $F^+$, in which the object is only partly covered by the inferred bounding box, are mostly examples as the carrot shown in Fig.~\ref{fig:frcnn_inference}, where the bounding box misses only a tiny fraction of the object. However, for many applications such as our use case of incremental learning of new objects, this small lack of impreciseness does not pose a problem since we observed that usually each part of an object is detected in a decent amount of training images such that in total all parts are seen by the model during training. On the other hand, we found in subsequent experiments that a slightly enlarged bounding box, as it is for instance the case for the potato in $F^+$ that is shown in Fig.~\ref{fig:frcnn_inference}, does not result in poor training results for the new object as long as tight bounding boxes are not an urgent requirement in the particular application.

The results for $I^+$ on test images of the iCubWorld data set are shown in Fig.~\ref{fig:icub_inference}. Again we found that the majority of objects is well detected in $I^+$ and the shift from $I$ to $I^+$ largely improves the  performance (see Fig.~\ref{fig:results_frcnn_more} in the appendix). We find that most objects are well detected even though they do not stand out clearly from the background (e.g.\ soda bottle in Fig.~\ref{fig:icub_inference}) or have not been in the training data set (e.g.\ remote in Fig.~\ref{fig:icub_inference}). A minority of the inferred bounding boxes does either cover the object only partly (again, as in the case of $F^+$, only a few pixels are missing), is too large due to inclusion of the human hand (e.g.\ hair clip)  or merge of correct with incorrectly detected boxes (e.g.\ perfume). 
Moreover, for both models $F$ and $I$, and thus also $F^+$ and $I^+$, it occurs only very rarely that an object is not detected at all.

\begin{figure*}[h!]
	\centering
	\includegraphics[width=0.9\textwidth]{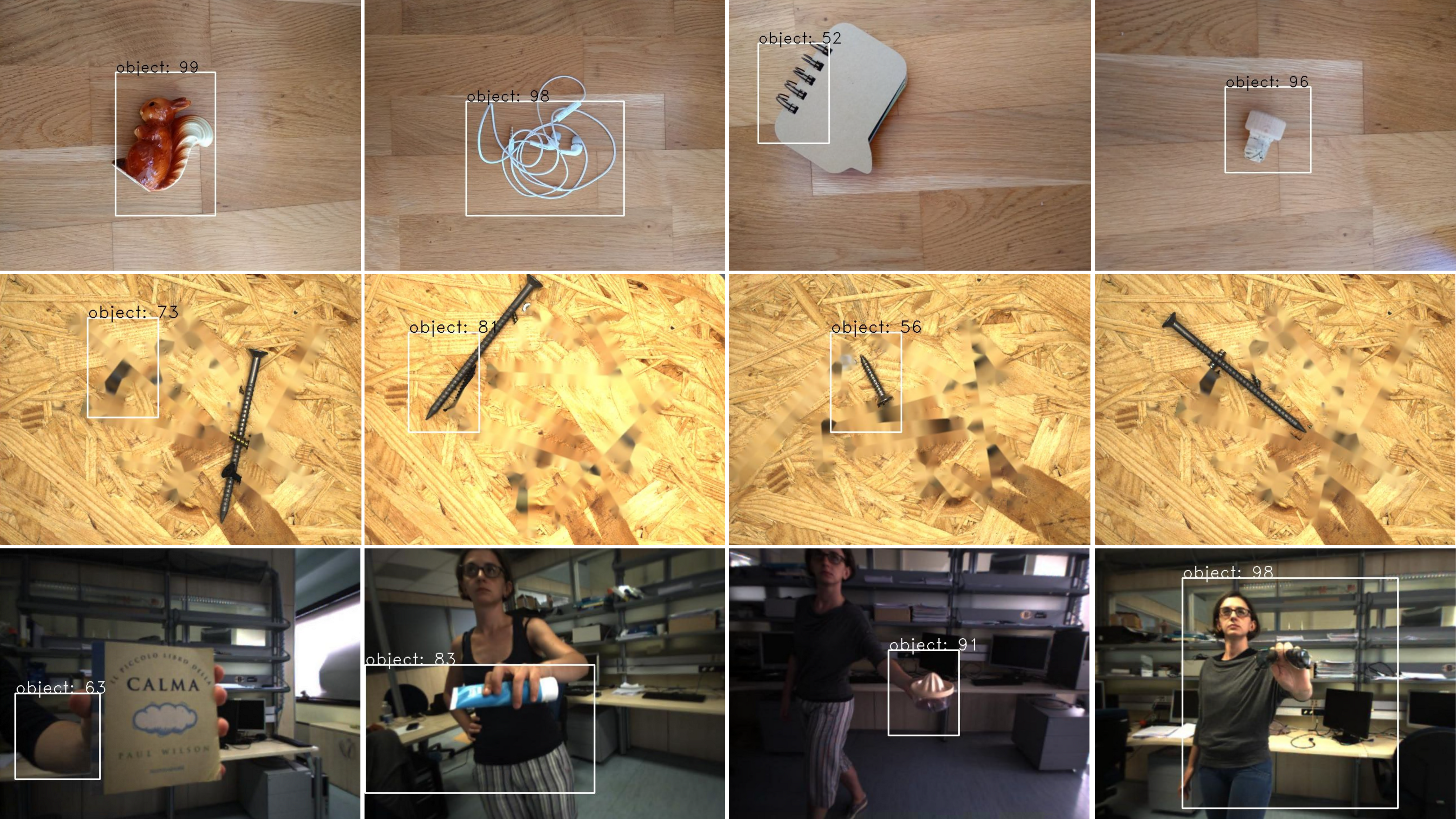}
	\captionsetup{singlelinecheck=false, width=0.9\linewidth}
	\caption{Inference results of the Faster R-CNN annotation model F$^+$ trained on the fruits data set for 200 epochs and applied to different data sets. The images in the first row are taken on a wooden floor, the second row contains images that have been modified from the original MVTec Screws data set \cite{ulrich2019comparison}. The last row shows examples from the iCubWorld data test.}
	\label{fig:fruits_others}
\end{figure*}

Finally, we tested the transferability of such annotation models to a different background. For this, we  applied the model $F^+$, that has been trained on fruits on a white background, to images containing various objects on different backgrounds. The results  are shown in Fig.~\ref{fig:fruits_others} for images on a wooden floor (taken by us), modified images from the MVTec Screws data set \cite{ulrich2019comparison}, and images from the iCubWorld data set. It can be seen that the model transfers only poorly to other data sets.  Applying $I^+$ to other data sets gives similar results on the MVTec data set, however, the model performs well on the wooden floor data set in 5 out of 6 images and mainly well on the fruits data set as well, where the object is mostly detected but sometimes with an inaccurate bounding box. Repeating these experiments with $F$ and $I$ results in many inaccurate bounding boxes as well as detection of multiple boxes.
In summary, these observations, together with the fairly good results of $I^+$ on the iCubWorld data set, suggest that the model explicitly \emph{learns} a given specific background but it can often be transferred to less complex backgrounds in combination with (P1) and (P2).

To conclude, we found that Faster R-CNN can be trained to distinguish (unknown) objects from a highly complex but homogeneous and specific background using  a relatively small amount of training data. The ability to transfer a trained model to a different data set is limited but works in some cases where the background is less complex than in the training data.

\subsubsection{Scaled Yolov4-p5}

\begin{figure*}[h!]
	\begin{minipage}[c]{1.0\linewidth}
		\subcaption[l]{\textbf{Metrics}}
		\centering
		\includegraphics[width=0.65\textwidth]{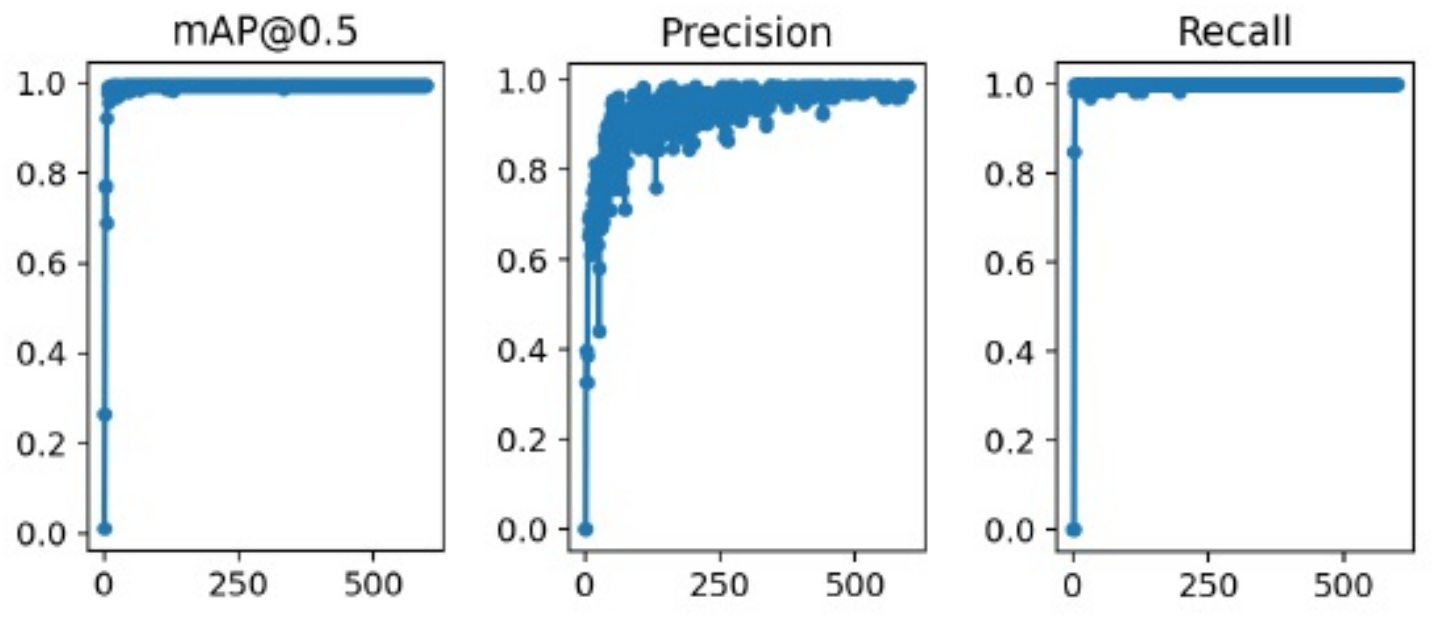}
	\end{minipage}
	\begin{minipage}[c]{1.0\linewidth}
		\vspace{0.2cm}
		\subcaption[l]{\textbf{Inference examples}}
		\centering
		\includegraphics[width=0.9\textwidth]{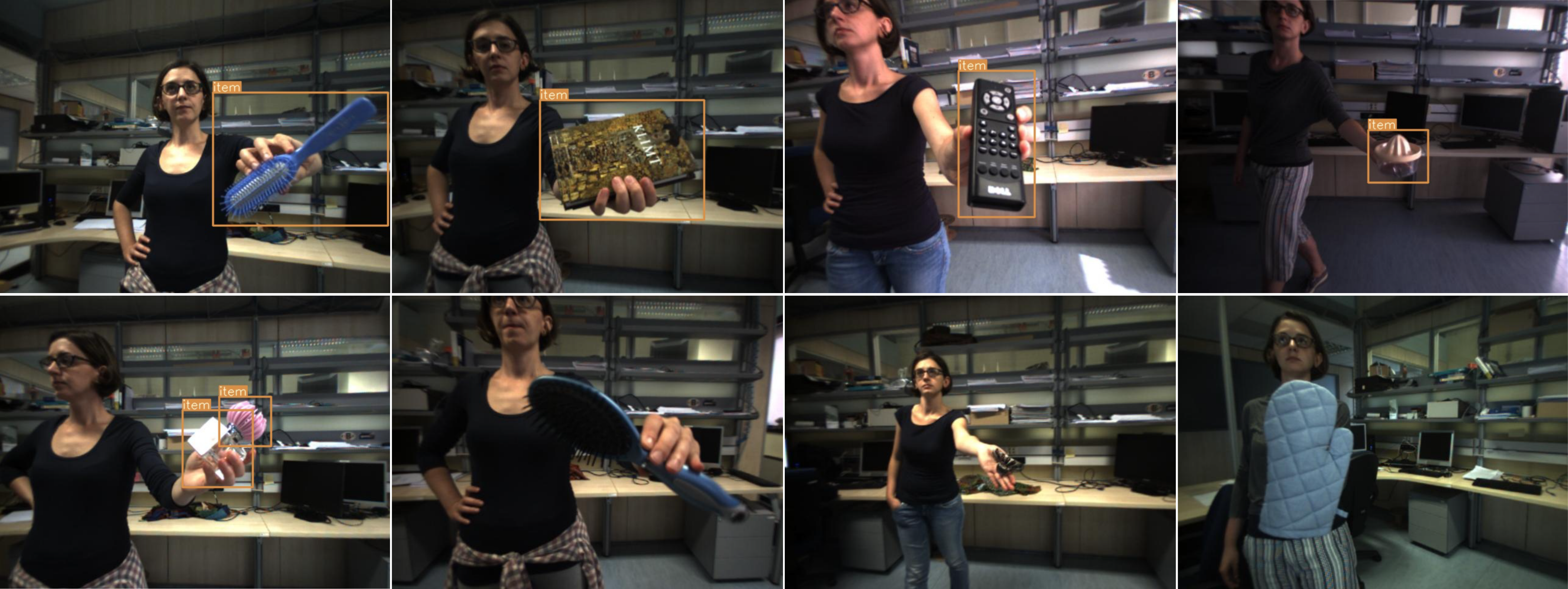}
	\end{minipage}
	\captionsetup{singlelinecheck=false, width=0.9\linewidth}
	\caption{Results on the iCubWorld data set using the Scaled Yolov4-p5 model. The model has been trained for 200 epochs.
			\textit{Panel (a):} The metrics (mAP@0.5, recall and precision) on the validation data sets are calculated after each epoch and reach high values after only a few epochs. For more details see Fig.~\ref{fig:metrics_icub_yolo}.\\
			\textit{Panel (b):} Inference examples on the iCubWorld test data set at epoch 175. Most objects are detected well, some are not detected at all (5 out of 82). In contrast to Faster R-CNN, there are only few cases where multiple bounding boxes have been detected (1 out of 82) and bounding boxes are more accurate with only rarely missing parts of the object (8 out of 82), even though (P1) and (P2) are not applied. 
	}
	\label{fig:results_icub_yolo_3}
\end{figure*}

Using a ScaledYolov4-p5 model architecture pretrained on COCO\footnote{\url{https://github.com/WongKinYiu/ScaledYOLOv4}}, we retrained the model on this iCubWorld data set. The results are shown in Fig.~\ref{fig:results_icub_yolo_3}.
As for the Faster R-CNN model, the metrics reach high values close to 1.0 after only a few epochs of training. In most cases, the model predicts good bounding boxes (see Fig.~\ref{fig:results_icub_yolo_3}(b), top row, for some examples). Poor results appear in cases where no object is detected or one object is detected as two objects. Overall in our test set with 82 images, only 5 objects were not detected and only once an object was detected as two objects. In contrast to the Faster R-CNN model, the bounding boxes are mostly very accurate although tight around the object (parts of the object have been missed in only 8 out of 82 test images). The additional post-processing steps (P1) and (P2) are therefore omitted. 

Finally, we tested the transferability of the model by applying it to data sets with different background. While the results on the MVTec data set are similarly bad as in the case of the Faster R-CNN model, the predictions on the fruits data set differ in comparison to $I^+$ (cf.\ Sec.~\ref{sec:results_frcnn}) in the sense that the bounding boxes are similarly accurate yet tighter for the Yolo model but, as already seen for the iCubWorld test data set, the Yolo model fails in detecting the object more often than Faster R-CNN. For the data set with the wooden floor, the Yolo model gives comparable good results as $I^+$ if (P1) is applied, otherwise only 3 out of 6 images are predicted correctly while in two cases the object is detected more than once. 

In summary, the ScaledYolov4-p5 architecture performs similarly well as the Faster R-CNN model in distinguishing (new) objects from homogeneous backgrounds. While the Yolo model fails in detecting the object more often than the Faster R-CNN model, its predicted bounding boxes are tighter and often more accurate. Moreover, the Yolo model does not require the application of the post-processing steps (P1) and (P2)  in most cases.

\section{Summary and Discussion}
\label{sec:summary}

In this work, we discussed how to adapt state-of-the-art object detection methods for the task of automatic bounding box annotation for the use case where the background is homogeneous and the object's label is provided by a human collaborator. 
An important advantage of using object detection methods instead of standard instance or semantic segmentation approaches is the usually smaller inference time, which is crucial for fast generation of training data. Moreover, in our pipeline, it has the added advantage that a single model architecture can be used for both tasks, bounding box annotation and object detection. In our experiments, we showed that both an adapted version of Faster R-CNN and the Scaled Yolov4-p5 architecture, can be trained to distinguish unknown objects from a complex but homogeneous background using only a small amount of training data. In contrast to Yolo, the Faster R-CNN strongly benefits from using two post-processing steps that merge multiple bounding boxes and enlarge the final box. On the other hand, Yolo fails more often in detecting objects than Faster R-CNN. Our results suggest that both models explicitly learn the specific background in the training data, which limits the ability to transfer a trained model to a data set with different background. However, it seems that models trained on more complex backgrounds can be transferred to data with less complex backgrounds. 
A further analysis of  the specific limits of this transferability  and to what extent they can be defined, would be an interesting question for future work. Moreover, it remains to further evaluate the minimum amount of required training data depending on the specific background as well as on transferability on different backgrounds.

\section*{Acknowledgements}
The research reported in this paper has been funded by the Federal Ministry for Climate Action, Environment, Energy, Mobility, Innovation and Technology (BMK), the Federal Ministry for Digital and Economic Affairs (BMDW), and the State of Upper Austria in the frame of the COMET - Competence Centers for Excellent Technologies Programme managed by Austrian Research Promotion Agency FFG.

\bibliographystyle{plain} 
\bibliography{references}

\begin{thebibliography}{10}

\bibitem{adhikari2021iterative}
Bishwo Adhikari and Heikki Huttunen.
\newblock Iterative bounding box annotation for object detection.
\newblock In {\em 2020 25th International Conference on Pattern Recognition
  (ICPR)}, pages 4040--4046. IEEE, 2021.

\bibitem{adhikari2018faster}
Bishwo Adhikari, Jukka Peltomaki, Jussi Puura, and Heikki Huttunen.
\newblock Faster bounding box annotation for object detection in indoor scenes.
\newblock In {\em 2018 7th European Workshop on Visual Information Processing
  (EUVIP)}, pages 1--6. IEEE, 2018.

\bibitem{apud2021automated}
Javier~Gibran Apud~Baca, Thomas Jantos, Mario Theuermann, Mohamed~Amin Hamdad,
  Jan Steinbrener, Stephan Weiss, Alexander Almer, and Roland Perko.
\newblock Automated data annotation for 6-dof ai-based navigation algorithm
  development.
\newblock {\em Journal of Imaging}, 7(11):236, 2021.

\bibitem{bargoti2017deep}
Suchet Bargoti and James Underwood.
\newblock Deep fruit detection in orchards.
\newblock In {\em 2017 IEEE International Conference on Robotics and Automation
  (ICRA)}, pages 3626--3633. IEEE, 2017.

\bibitem{bendale2015towards}
Abhijit Bendale and Terrance Boult.
\newblock Towards open world recognition.
\newblock In {\em Proceedings of the IEEE conference on computer vision and
  pattern recognition}, pages 1893--1902, 2015.

\bibitem{bendale2016towards}
Abhijit Bendale and Terrance~E Boult.
\newblock Towards open set deep networks.
\newblock In {\em Proceedings of the IEEE conference on computer vision and
  pattern recognition}, pages 1563--1572, 2016.

\bibitem{bochkovskiy2020yolov4}
Alexey Bochkovskiy, Chien-Yao Wang, and Hong-Yuan~Mark Liao.
\newblock Yolov4: Optimal speed and accuracy of object detection.
\newblock {\em arXiv preprint arXiv:2004.10934}, 2020.

\bibitem{cheng2018survey}
Qimin Cheng, Qian Zhang, Peng Fu, Conghuan Tu, and Sen Li.
\newblock A survey and analysis on automatic image annotation.
\newblock {\em Pattern Recognition}, 79:242--259, 2018.

\bibitem{everingham2010pascal}
Mark Everingham, Luc Van~Gool, Christopher~KI Williams, John Winn, and Andrew
  Zisserman.
\newblock The pascal visual object classes (voc) challenge.
\newblock {\em International journal of computer vision}, 88(2):303--338, 2010.

\bibitem{fanello2013icub}
Sean Fanello, Carlo Ciliberto, Matteo Santoro, Lorenzo Natale, Giorgio Metta,
  Lorenzo Rosasco, and Francesca Odone.
\newblock icub world: Friendly robots help building good vision data-sets.
\newblock In {\em Proceedings of the IEEE Conference on Computer Vision and
  Pattern Recognition Workshops}, pages 700--705, 2013.

\bibitem{ge2020towards}
Ce~Ge, Jing Wang, Jingyu Wang, Qi~Qi, Haifeng Sun, and Jianxin Liao.
\newblock Towards automatic visual inspection: A weakly supervised learning
  method for industrial applicable object detection.
\newblock {\em Computers in Industry}, 121:103232, 2020.

\bibitem{yolox2021}
Zheng Ge, Songtao Liu, Feng Wang, Zeming Li, and Jian Sun.
\newblock {YOLOX:} exceeding {YOLO} series in 2021.
\newblock {\em arXiv preprint arXiv:2107.08430}, 2021.

\bibitem{geissdexa2022}
Manuela Gei{\ss}, Martin Baresch, Georgios Chasparis, Edwin Schweiger, Nico
  Teringl, and Michael Zwick.
\newblock Fast and automatic object registration for human-robot collaboration
  in industrial manufacturing.
\newblock {\em arXiv preprint arXiv:2204.00597}, 2022.

\bibitem{He_2017_ICCV}
Kaiming He, Georgia Gkioxari, Piotr Dollar, and Ross Girshick.
\newblock Mask r-cnn.
\newblock In {\em Proceedings of the IEEE International Conference on Computer
  Vision (ICCV)}, Oct 2017.

\bibitem{HE2021106622}
Xin He, Kaiyong Zhao, and Xiaowen Chu.
\newblock Automl: A survey of the state-of-the-art.
\newblock {\em Knowledge-Based Systems}, 212:106622, 2021.

\bibitem{huang2017speed}
Jonathan Huang, Vivek Rathod, Chen Sun, Menglong Zhu, Anoop Korattikara,
  Alireza Fathi, Ian Fischer, Zbigniew Wojna, Yang Song, Sergio Guadarrama,
  et~al.
\newblock Speed/accuracy trade-offs for modern convolutional object detectors.
\newblock In {\em Proceedings of the IEEE conference on computer vision and
  pattern recognition}, pages 7310--7311, 2017.

\bibitem{jiang2017face}
Huaizu Jiang and Erik Learned-Miller.
\newblock Face detection with the faster r-cnn.
\newblock In {\em 2017 12th IEEE international conference on automatic face \&
  gesture recognition (FG 2017)}, pages 650--657. IEEE, 2017.

\bibitem{kim2020comparison}
Jeong-ah Kim, Ju-Yeong Sung, and Se-ho Park.
\newblock Comparison of faster-rcnn, yolo, and ssd for real-time vehicle type
  recognition.
\newblock In {\em 2020 IEEE International Conference on Consumer
  Electronics-Asia (ICCE-Asia)}, pages 1--4. IEEE, 2020.

\bibitem{kiyokawa2019fully}
Takuya Kiyokawa, Keita Tomochika, Jun Takamatsu, and Tsukasa Ogasawara.
\newblock Fully automated annotation with noise-masked visual markers for
  deep-learning-based object detection.
\newblock {\em IEEE Robotics and Automation Letters}, 4(2):1972--1977, 2019.

\bibitem{konyushkova2018learning}
Ksenia Konyushkova, Jasper Uijlings, Christoph~H Lampert, and Vittorio Ferrari.
\newblock Learning intelligent dialogs for bounding box annotation.
\newblock In {\em Proceedings of the IEEE Conference on Computer Vision and
  Pattern Recognition}, pages 9175--9184, 2018.

\bibitem{krizhevsky2012imagenet}
Alex Krizhevsky, Ilya Sutskever, and Geoffrey~E Hinton.
\newblock Imagenet classification with deep convolutional neural networks.
\newblock {\em Advances in neural information processing systems},
  25:1097--1105, 2012.

\bibitem{le2020toward}
Trung-Nghia Le, Akihiro Sugimoto, Shintaro Ono, and Hiroshi Kawasaki.
\newblock Toward interactive self-annotation for video object bounding box:
  Recurrent self-learning and hierarchical annotation based framework.
\newblock In {\em Proceedings of the IEEE/CVF Winter Conference on Applications
  of Computer Vision}, pages 3231--3240, 2020.

\bibitem{lecun2015deep}
Yann LeCun, Yoshua Bengio, and Geoffrey Hinton.
\newblock Deep learning.
\newblock {\em nature}, 521(7553):436--444, 2015.

\bibitem{Lee2020PitfallsAP}
Chia-Yen Lee and Chen~Fu Chien.
\newblock Pitfalls and protocols of data science in manufacturing practice.
\newblock {\em Journal of Intelligent Manufacturing}, pages 1--19, 2020.

\bibitem{Lin_2017_CVPR}
Tsung-Yi Lin, Piotr Dollar, Ross Girshick, Kaiming He, Bharath Hariharan, and
  Serge Belongie.
\newblock Feature pyramid networks for object detection.
\newblock In {\em Proceedings of the IEEE Conference on Computer Vision and
  Pattern Recognition (CVPR)}, July 2017.

\bibitem{lin2017focal}
Tsung-Yi Lin, Priya Goyal, Ross Girshick, Kaiming He, and Piotr Doll{\'a}r.
\newblock Focal loss for dense object detection.
\newblock In {\em Proceedings of the IEEE international conference on computer
  vision}, pages 2980--2988, 2017.

\bibitem{lin2014microsoft}
Tsung-Yi Lin, Michael Maire, Serge Belongie, James Hays, Pietro Perona, Deva
  Ramanan, Piotr Doll{\'a}r, and C~Lawrence Zitnick.
\newblock Microsoft coco: Common objects in context.
\newblock In {\em European conference on computer vision}, pages 740--755.
  Springer, 2014.

\bibitem{Wei2016}
Wei Liu, Dragomir Anguelov, Dumitru Erhan, Christian Szegedy, Scott Reed,
  Cheng-Yang Fu, and Alexander~C. Berg.
\newblock Ssd: Single shot multibox detector.
\newblock In Bastian Leibe, Jiri Matas, Nicu Sebe, and Max Welling, editors,
  {\em Computer Vision -- ECCV 2016}, pages 21--37, Cham, 2016. Springer
  International Publishing.

\bibitem{matheson2019Cobot}
Eloise Matheson, Riccardo Minto, Emanuele Zampieri, Maurizio Faccio, and Giulio
  Rosati.
\newblock Human–robot collaboration in manufacturing applications: A review.
\newblock {\em Robotics}, 8:100, 12 2019.

\bibitem{Mehrabi2000ReconfManufacturing}
M.~Mehrabi, A.~Ulsoy, and Yoram Koren.
\newblock Reconfigurable manufacturing systems: Key to future manufacturing.
\newblock {\em Journal of Intelligent Manufacturing}, 11, 08 2000.

\bibitem{papadopoulos2016we}
Dim~P Papadopoulos, Jasper~RR Uijlings, Frank Keller, and Vittorio Ferrari.
\newblock We don't need no bounding-boxes: Training object class detectors
  using only human verification.
\newblock In {\em Proceedings of the IEEE conference on computer vision and
  pattern recognition}, pages 854--863, 2016.

\bibitem{Redmon_2016_CVPR}
Joseph Redmon, Santosh Divvala, Ross Girshick, and Ali Farhadi.
\newblock You only look once: Unified, real-time object detection.
\newblock In {\em Proceedings of the IEEE Conference on Computer Vision and
  Pattern Recognition (CVPR)}, June 2016.

\bibitem{10.1145/3472291}
Pengzhen Ren, Yun Xiao, Xiaojun Chang, Po-Yao Huang, Zhihui Li, Brij~B. Gupta,
  Xiaojiang Chen, and Xin Wang.
\newblock A survey of deep active learning.
\newblock {\em ACM Comput. Surv.}, 54(9), October 2021.

\bibitem{ren2015faster}
Shaoqing Ren, Kaiming He, Ross Girshick, and Jian Sun.
\newblock Faster r-cnn: Towards real-time object detection with region proposal
  networks.
\newblock {\em Advances in neural information processing systems}, 28:91--99,
  2015.

\bibitem{russakovsky2015imagenet}
Olga Russakovsky, Jia Deng, Hao Su, Jonathan Krause, Sanjeev Satheesh, Sean Ma,
  Zhiheng Huang, Andrej Karpathy, Aditya Khosla, Michael Bernstein, et~al.
\newblock Imagenet large scale visual recognition challenge.
\newblock {\em International journal of computer vision}, 115(3):211--252,
  2015.

\bibitem{scheirer2012toward}
Walter~J Scheirer, Anderson de~Rezende~Rocha, Archana Sapkota, and Terrance~E
  Boult.
\newblock Toward open set recognition.
\newblock {\em IEEE transactions on pattern analysis and machine intelligence},
  35(7):1757--1772, 2012.

\bibitem{Simon_2019_CVPR_Workshops}
Martin Simon, Karl Amende, Andrea Kraus, Jens Honer, Timo Samann, Hauke
  Kaulbersch, Stefan Milz, and Horst Michael~Gross.
\newblock Complexer-yolo: Real-time 3d object detection and tracking on
  semantic point clouds.
\newblock In {\em Proceedings of the IEEE/CVF Conference on Computer Vision and
  Pattern Recognition (CVPR) Workshops}, June 2019.

\bibitem{ulrich2019comparison}
Markus Ulrich, Patrick Follmann, and Jan-Hendrik Neudeck.
\newblock A comparison of shape-based matching with deep-learning-based object
  detection.
\newblock {\em tm-Technisches Messen}, 86(11):685--698, 2019.

\bibitem{wang2021scaled}
Chien-Yao Wang, Alexey Bochkovskiy, and Hong-Yuan~Mark Liao.
\newblock Scaled-yolov4: Scaling cross stage partial network.
\newblock In {\em Proceedings of the IEEE/CVF Conference on Computer Vision and
  Pattern Recognition}, pages 13029--13038, 2021.

\bibitem{wong2019assistive}
Vivian Wen~Hui Wong, Max Ferguson, Kincho~H Law, and Yung-Tsun~Tina Lee.
\newblock An assistive learning workflow on annotating images for object
  detection.
\newblock In {\em 2019 IEEE International Conference on Big Data (Big Data)},
  pages 1962--1970. IEEE, 2019.

\bibitem{wu2020automatic}
Joy Wu, Yaniv Gur, Alexandros Karargyris, Ali~Bin Syed, Orest Boyko, Mehdi
  Moradi, and Tanveer Syeda-Mahmood.
\newblock Automatic bounding box annotation of chest x-ray data for
  localization of abnormalities.
\newblock In {\em 2020 IEEE 17th international symposium on biomedical imaging
  (ISBI)}, pages 799--803. IEEE, 2020.

\bibitem{Zhao2019}
Zhong-Qiu Zhao, Peng Zheng, Shou-Tao Xu, and Xindong Wu.
\newblock Object detection with deep learning: A review.
\newblock {\em IEEE Transactions on Neural Networks and Learning Systems},
  30(11):3212--3232, 2019.

\end{thebibliography}

\begin{appendix}
	
\section{More Results}
In this section, we present more detailed results about the experiments of Sec.~\ref{sec:experiments}.
	
\begin{figure*}[htb]
	\begin{minipage}[c]{1.0\linewidth}
		\subcaption[l]{\textbf{Metrics \& Loss}}
		\centering
		\includegraphics[width=0.9\textwidth]{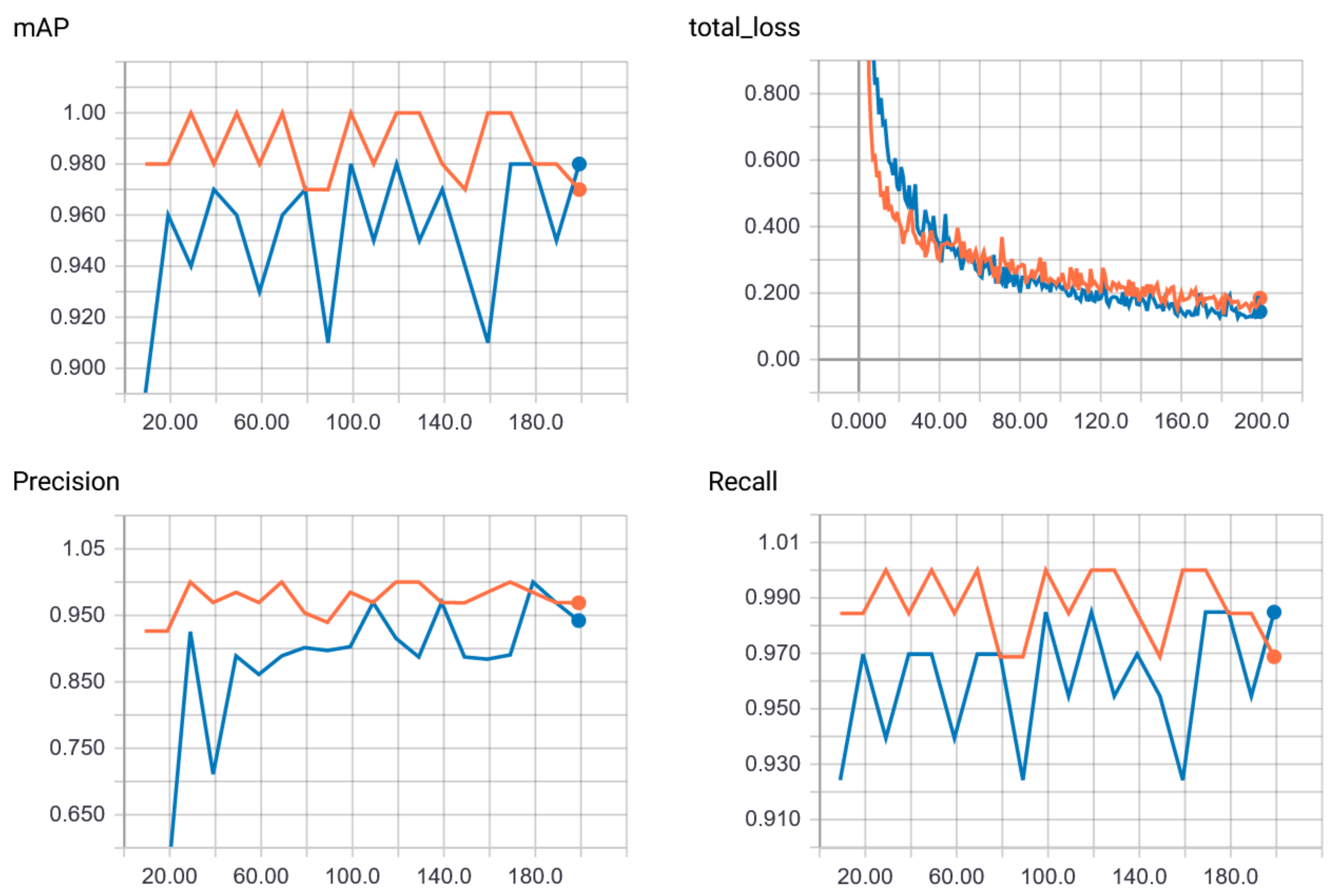}
	\end{minipage} 
	
	\begin{minipage}[c]{1.0\linewidth}
		\vspace{0.2cm}
		\subcaption{\textbf{Inference Results on Test Data}}
		\begin{center}
		\resizebox{0.9\textwidth}{!}{%
			\begin{tabular}[b]{|l|cccc|cccc|cccc|cccc|cccc|}
				\hline
				& \multicolumn{4}{c}{Epoch 10} & \multicolumn{4}{c}{Epoch 50} & \multicolumn{4}{c}{Epoch 100} & \multicolumn{4}{c}{Epoch 150} & \multicolumn{4}{c}{Epoch 200}\\
				& $F$ & $F^+$ & $I$ & $I^+$ & $F$ & $F^+$ & $I$ & $I^+$ & $F$ & $F^+$ & I & $I^+$ & $F$ & $ F^+$ & $I$ & $I^+$ & $F$ & $F^+$ & $I$ & $I^+$\\
				\hline
				correct& 32 & 84& 10& 17 & 43 & 83 &37 & 67 & 58 & 92 &36 & 62& 48 &85 &51 &66 & 51&87 &50 &70\\
				not detected &  0 &0 & 0 & 0 & 0 &0 & 1 &1 & 0 &0 & 0 &0 & 0 &0 & 1 &1 & 1&1 &1 &1\\ 
				multiple times &  14 & - &6 &- & 11 & -&1 &- & 16 & - &3 & -& 14 & -&0 & -& 14& -&0 &-\\
				partly by single bounding box &  54 &16 &33 & 2& 46 & 17&38 &5 & 26 & 8&32 & 6& 38 &15 & 27& 8& 34& 12&24 &4\\
				correct + other parts &  0 & 0& 31& 63&0 & 0&5 & 9& 0 & 0&10 & 14& 0 & 0&4 &7 & 0& 0& 6&7\\
				background &  0 &0 & 21 & - & 0 &0 & 2 & - & 0 &0 & 4 &- & 0 &0 & 3 &- & 0&0 & 4 &-\\
				head/face, hand, body & - &- & 42, 7, 6 & -& - &- & 1, 6, 2 & -& - &- & 4, 10, 2 &- & - &- &  1, 3, 1&- & -&- & 1, 2, 0  &-\\
				\hline
		\end{tabular}
	}
	\end{center}
	\end{minipage}
	\captionsetup{singlelinecheck=false, width=0.9\linewidth}
	\caption{Results on the fruits data set and the iCubWorld data set using Faster R-CNN. For both data sets, the Faster R-CNN architecture has been trained for 200 epochs.\\
			\textit{Panel (a):}  The metrics (mAP@0.5, recall and precision) on the validation data sets are calculated after each 10th epoch (orange: fruits, blue: iCubWorld). For both data sets, the metrics reach high values after only a few epochs. The loss for the training data set is shown after each epoch.\\
			\textit{Panel (b):} Inference results on the test data sets of 100 (fruits) resp.\ 82 (iCubWorld) images, both containing more object classes than the training data. For more details about the different categories see the text.}
	\label{fig:results_frcnn_more}
\end{figure*}	
	
Fig.~\ref{fig:results_frcnn_more} shows additional results for the experiments with the Faster R-CNN architecture. In addition to the metrics that have already been shown in Fig.~\ref{fig:results_frcnn}, it also contains the total training loss. Moreover, detailed inference results for the models $F$ and $F^+$ on the fruits test data set and for $I$ and $I^+$ on the iCubWorld test data set are shown for different epochs from 10 to 200. The results are assigned to 7 different classes:\\
\textit{correct:} Good detection, i.e., correct label, good bounding box and no false positives/negatives.\\
\textit{not detected:} The object is not detected.\\
\textit{multiple times:} Multiple bounding boxes covering the object or parts of it are inferred.\\
\textit{partly by single bounding box:} The object is detected by a single bounding box but the box misses parts of the object. We explicitly emphasize that this mostly corresponds to only very small missing parts (see also Sec.~\ref{sec:results_frcnn}). \\
\textit{correct + other parts:} The object is well detected but other parts of the image are detected as well (see also the remaining two categories). We also count here objects that are detected by a bounding box that is at least twice as large as necessary, containing also e.g. the person's hand in the iCubWorld data set.\\
\textit{background:} The background is detected. For the iCubWorld data set, we do not include here the detection of (parts of) the person presenting the object. This is covered by the next category.\\
\textit{head/face, hand, body:} The head/face, hand, or other parts of the body of the person presenting the object are detected (only applies to the iCubWorld data set).\\

\begin{figure*}[htb]
	\centering
	\includegraphics[width=0.9\textwidth]{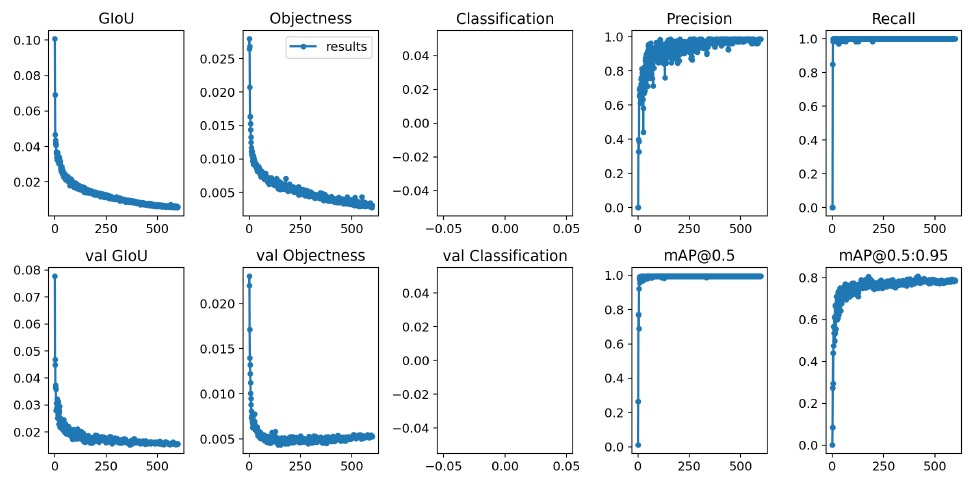}
	\captionsetup{singlelinecheck=false, width=0.9\linewidth}
	\caption{Training losses and metrics for the Scaled Yolov4-p5 trained on the iCubWorld data set (see Sec.~\ref{sec:data}) for 200 epochs.}
	\label{fig:metrics_icub_yolo}
\end{figure*}

For the experiment with the Scaled Yolov4-p5, the complete set of metrics and losses is shown in Fig.~\ref{fig:metrics_icub_yolo}.

\end{appendix}

\end{document}